\documentclass[10pt, journal]{ieeeconf}
\IEEEoverridecommandlockouts

\usepackage{cite}
\usepackage{amsmath,amssymb,amsfonts}
\usepackage{algorithmic}
\usepackage{graphicx}
\usepackage{subcaption}
\usepackage{textcomp}
\usepackage{multirow}
\usepackage{xcolor}
\usepackage{color, colortbl}
\definecolor{Gray}{gray}{0.9}
\usepackage{bm}
\usepackage{booktabs}
\usepackage{svg}
\svgsetup{inkscapelatex=false}
\usepackage{tikz}
\usepackage{hyperref}
\usepackage{cleveref}
\usepackage[T1]{fontenc}

\def\BibTeX{{\rm B\kern-.05em{\sc i\kern-.025em b}\kern-.08em
    T\kern-.1667em\lower.7ex\hbox{E}\kern-.125emX}}
\usetikzlibrary{arrows,decorations.pathmorphing,backgrounds,fit,positioning,calc,shapes}

\usepackage[linesnumbered,ruled,algo2e,resetcount]{algorithm2e}

\begin{document}

\title{The First WARA Robotics Mobile Manipulation Challenge - \\ Lessons Learned}

\author{\authorblockN{
David Cáceres Domínguez$^{a*}$,
Marco Iannotta$^{a*}$,
Abhishek Kashyap$^{a*}$,
Shuo Sun$^{a*}$,
Yuxuan Yang$^{a*}$,
Christian Cella$^{b*}$,
Matteo Colombo$^{b*}$,
Martina Pelosi$^{b*}$,
Giuseppe F. Preziosa$^{b*}$,
Alessandra Tafuro$^{b*}$,
Isacco Zappa$^{b*}$,
Finn Busch$^{c*}$,
Yifei Dong$^{c*}$,
Alberta Longhini$^{c*}$,
Haofei Lu$^{c*}$,
Rafael I. Cabral Muchacho$^{c*}$,
Jonathan Styrud$^{c*}$,
Sebastiano Fregnan$^{d*}$,
Marko Guberina$^{d*}$,
Zheng Jia$^{d*}$,
Graziano Carriero$^{d,e*}$, 
Sofia Lindqvist$^{f}$,
Silvio Di Castro$^{f}$,
Matteo Iovino$^{g}$}
\thanks{
The WASP Research Arena (WARA)-Robotics, hosted by ABB Corporate Research Center in Västerås, Sweden is financially supported by the Wallenberg AI, Autonomous Systems, and Software Program (WASP) funded by the Knut and Alice Wallenberg Foundation.}
\thanks{$^{*}$These authors have contributed equally to this work.}
\thanks{$^{a}$Center for Applied Autonomous Sensor Systems (AASS), Örebro University, Örebro, Sweden}
\thanks{$^{b}$Politecnico di Milano, Dipartimento di Elettronica, Informazione e Bioingegneria, Milano, Italy}
\thanks{$^{c}$Division of Robotics, Perception and Learning, Royal Institute of Technology (KTH), Stockholm, Sweden}
\thanks{$^{d}$Lund Technical University (LTH), Lund, Sweden}
\thanks{$^{e}$University of Basilicata, Italy}
\thanks{$^{f}$Pharmaceutical Technology \& Development, R\&D, AstraZeneca, Gothenburg, Sweden}
\thanks{$^{g}$ABB Corporate Research, Västerås, Sweden}
}

\maketitle
\begin{abstract}

The first WARA Robotics Mobile Manipulation Challenge, held in December 2024 at ABB Corporate Research in Västerås, Sweden, addressed the automation of task-intensive and repetitive manual labor in laboratory environments --- specifically the transport and cleaning of glassware. Designed in collaboration with AstraZeneca, the challenge invited academic teams to develop autonomous robotic systems capable of navigating human-populated lab spaces and performing complex manipulation tasks, such as loading items into industrial dishwashers. This paper presents an overview of the challenge setup, its industrial motivation, and the four distinct approaches proposed by the participating teams. 
We summarize lessons learned from this edition and propose improvements in design to enable a more effective second iteration to take place in 2025. 
The initiative bridges an important gap in effective academia-industry collaboration within the domain of autonomous mobile manipulation systems by promoting the development and deployment of applied robotic solutions in real-world laboratory contexts.

\end{abstract}


\section{Introduction}
Modern biomedical research often relies on precise and repetitive lab procedures carried out by highly qualified personnel. Despite advancements in experimental techniques and research technologies, routine tasks --- such as preparing solutions, cleaning instruments, or organizing equipment --- continue to be performed manually in many laboratories. In particular, washing glassware remains a labor-intensive process that, while crucial for ensuring cleanliness and adherence to quality standards, does not require specialized expertise. Instead, it consumes the valuable time of researchers who could otherwise devote their skills to scientific duties.

\begin{figure}[t!]
    \centering
    \includegraphics[width=.9\columnwidth]{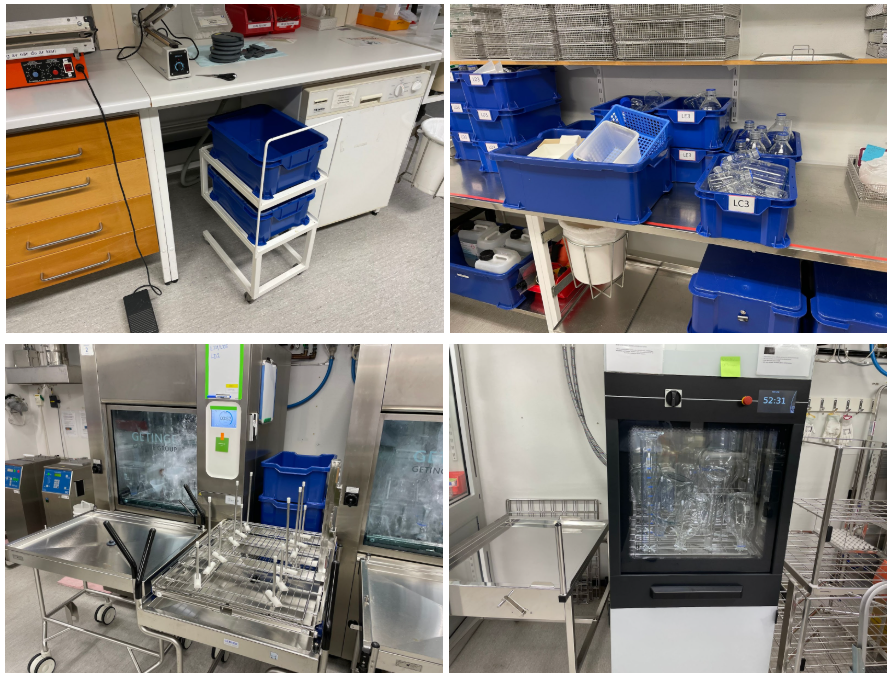}
    \caption{Lab Automation use-case in the AstraZeneca chemical laboratories: a human operator drives a cart around the lab to collect dirty glassware, organizes it in a dishwasher tray, and finally inserts the tray in the dishwasher.}
    \label{fig:az_lab}
\end{figure}

The primary motivation behind the first WARA Robotics\footnote{\url{https://wara-robotics.se/}} Mobile Manipulation Challenge, hosted at ABB Corporate Research in Västerås, Sweden, was to address the problem of task-intensive lab work. The challenge aimed to push the boundaries of mobile manipulation research by tackling a highly relevant but under-explored real-world use case: automating routine laboratory workflows. Specifically, it aimed to develop systems capable of autonomously navigating human-populated spaces to transport carts of glassware from one location to another and manipulate items for loading into an industrial dishwasher. The challenge use-case was proposed by AstraZeneca (as depicted in~\Cref{fig:az_lab}) and it encompassed multiple interconnected sub-tasks, including safe navigation, grasping and dexterous manipulation, as well as perception for object detection and pose estimation. By tackling these complex research areas, participating teams created innovative robotic solutions that have the potential to transform lab routines, thereby allowing scientific staff to reallocate their time to high-value-adding research activities.

Engaging in this type of challenge enables WASP industrial partners to guide the direction of academic groups, ensuring alignment with practical, real-world applications. This strategic collaboration accelerates progress in the field while providing mutual benefits to all parties involved, particularly in learning and knowledge development. While the specifics of these benefits cannot be precisely predicted, it is expected that they will emerge in the future, contributing to the ongoing growth and advancement of the robotics community.

This paper describes the four different approaches to solve the task as proposed by the participating teams and outlines the results and lessons learned. Importantly, the challenge also served as a way to evaluate the technology readiness of academic research, offering a realistic testbed for translating theoretical advances into deployable systems. These insights will help formulating the $2^{nd}$ iteration of the challenge.

\section{Challenge Rules}\label{sec:rules}

The participating teams were challenged to develop a system that autonomously and safely navigates in a human-populated lab environment, localizes a cart loaded with glassware that needs washing, transports the cart to a designated dishwasher room, and manipulates the glassware for loading into an industrial dishwasher. There were no restrictions on the choice of the robotic platform nor the solution strategy.

\begin{figure}[t]
    \centering
    \includegraphics[width=.8\columnwidth]{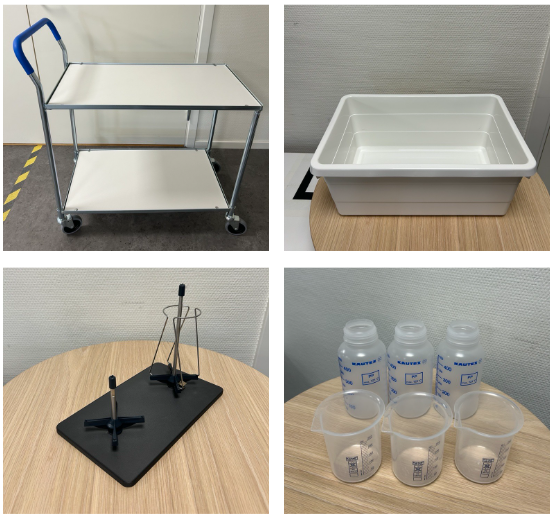}
    \caption{All the teams received equipment to mock-up the lab automation use-case. The equipment included a cart with 4 castor wheels, a plastic bin, a plastic base hosting 2 different types of dishwasher pins (to mimic the dishwasher tray), 3 plastic bottles, and 3 plastic beakers.}
    \label{fig:mockup}
\end{figure}

A challenge mock-up kit (shown in~\Cref{fig:mockup}) was sent to the teams to allow them to develop and test their own solutions before coming to the WARA Robotics lab in Västerås. In the mock-up kit, the lab glassware was replaced by plastic items, comprising 3 bottles and 3 beakers.

The challenge was composed of the following sub-tasks.

\subsection{Carting glassware that needs washing}\label{subtaskA}
The first sub-task concerned a mobile robot navigation and manipulation setup. In this task, the robot had to navigate an environment to a preset goal location where a cart had been positioned. A bin filled with randomly placed plasticware was positioned on the cart. Then, the robot had to pull/push the cart and navigate again to a second marked location in a room where the dishwasher tray was set up. It was assumed that there were no closed doors for the robot to manipulate along the way. For this task the robot had only onboard sensors available which included for instance lidars and cameras. WARA Robotics provided a 2D occupancy map of the environment as prior and allowed teams to run a (teleoperated) mapping session prior to addressing the challenge, if desired. The task was considered successfully completed if the robot was able to move the cart from the start to the goal location. It was also expected that the path the cart needs to take included at least one left and at least one right turn. 

\subsection{Manipulating glassware and dishwasher}\label{subtaskB}
The second sub-task assumed that the robot was already at the dishwasher tray location and was presented with a table-top scenario. A bin filled with randomly placed plasticware was provided at one end of the table and the dishwasher tray at the other end, with both placed roughly within predefined workspace zones. The task of the robot was then to select items from the bin, to pick them up, and to safely insert them onto the pins sticking out from the dishwasher tray. Different levels of difficulty were envisioned for this task, ranging from very easy (small workspace feasible for a fixed-base robot, homogeneous non-transparent items tagged with QR codes) to very hard (large workspace that requires a mobile base, heterogeneous transparent plastic objects). This task focused on random bin picking and automatic loading, while unloading the tray was still performed manually.

\section{Approaches}

This section provides an overview of each team's approach to solve the proposed challenge. Four teams from different universities took part in the challenge: Örebro University (ÖrU), Politecnico di Milano (PoliMi), KTH Royal Institute of Technology (KTH), and Lund University (LTH). A short overview of the strategy used by each team and what task they focused on is reported in Table~\ref{table:teams_overview}, where Task~A represents the mobile navigation sub-task (\ref{subtaskA}), while Task~B indicates the dishwasher loading sub-task (\ref{subtaskB}).

\begin{table}[ht]
\centering
\scriptsize
\caption{Summary of the strategy adopted by each team.}
\begin{tabular}{c c c c c}
\toprule
\textbf{Team} & \textbf{Platform}     & \textbf{Gripper} & \textbf{Perception} & \textbf{Task} \\
\midrule
ÖrU & Franka Panda & parallel & RGB-D camera &  B \\
\midrule
PoliMi  & GoFa~5  & parallel + suction  & RGB-D camera & B \\
\midrule
KTH   &  Mobile YuMi   & parallel & RGB-D camera &  A\&B  \\
\midrule
LTH   &  Mobile YuMi  & parallel & RGB-D camera & A  \\
\bottomrule
\end{tabular}
\label{table:teams_overview}
\end{table}

\subsection{\"Orebro University (ÖrU)}
\begin{figure}[ht]
    \centering
    \begin{subfigure}[b]{0.35\columnwidth}
        \includegraphics[width=\linewidth]{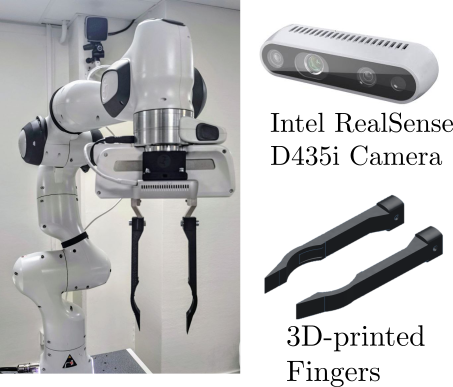}
        \caption{}
        \label{fig:oru_setup}
    \end{subfigure}
    \hfill
    \begin{subfigure}[b]{0.63\columnwidth}
        \includegraphics[width=\linewidth]{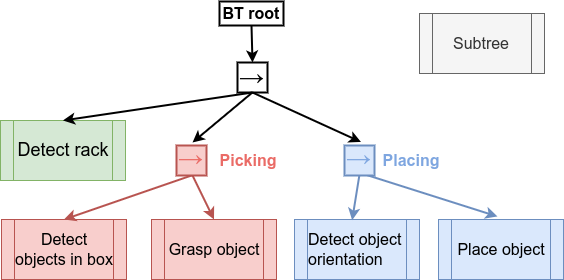}
        \caption{}
        \label{fig:oru_bt}
    \end{subfigure}
    \caption{(a) Robot hardware setup. (b) Behavior Tree.}
\end{figure}

The ÖrU team focused on subtask B, dishwasher loading. 
The robotic manipulation system consisted of a Franka Emika Panda arm with custom 3D-printed fingers and an Intel RealSense D435i camera, as shown in~\Cref{fig:oru_setup}. Together, the vision, grasping, and task execution systems enabled the robot to detect, grasp, and place transparent bottles and beakers onto designated dishwasher pins.

\subsubsection{Vision System}
The vision system relied on the RGB-D camera to detect and localize objects and pins. It needed to overcome challenges inherent to transparent surfaces, such as unreliable depth data and noisy point clouds.

\begin{figure}[t]
    \centering
    
    \begin{subfigure}[b]{\linewidth}
        \centering
        \includegraphics[width=\linewidth]{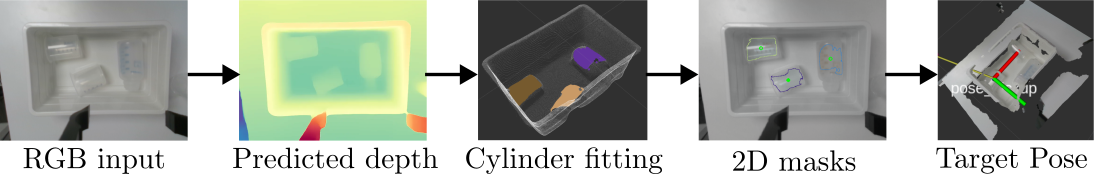}
        \caption{}
        \label{fig:oru_bottlebeaker_detection}
    \end{subfigure}
    
    \begin{subfigure}[b]{\linewidth}
        \centering
        \includegraphics[width=\linewidth]{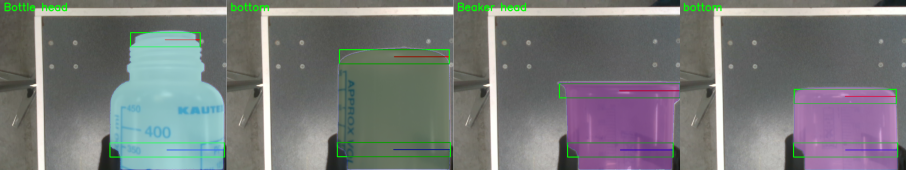}
        \caption{}
        \label{fig:oru_opening_detection}
    \end{subfigure}


    \caption{(a) Bottle and beaker detection and localization. (b) Opening direction detection. }
    \label{fig:oru_perception}
\end{figure}

    \paragraph{Bottle and Beaker Detection and Localization}\label{par:oru_detection}
    To overcome the problem of poor depth information, they employed MoGe~\cite{wang2024moge} for monocular depth estimation from RGB images which produced more refined depth information compared to the camera depth image. The point cloud from estimated depth was used to segment the bin with the objects in it. They then applied cylinder fitting to the points inside the bin to detect the cylindrically shaped objects, mapping the inliers from the fitting process to pixels in the depth image. Using an empirically developed ranking system based on pixel count (indicating visibility) and aspect ratio (closeness to 1 indicating better geometry) that ensured robust object selection for manipulation, the best detection was selected for grasping. The process overview is shown in~\Cref{fig:oru_bottlebeaker_detection}. The estimated depth is relative, not metric, but with a single object layer and near-vertical camera it was converted to Cartesian coordinates using the camera's intrinsic parameters. 

    \paragraph{Opening Direction Detection} \label{par:oru_opening}
    Cylinder fitting locates bottles and beakers but doesn't reveal the openings' orientation. To ensure proper placement with the opening facing downward, an extra step was added to identify its direction. Once the object is held in the gripper, they segmented it with SAM2~\cite{ravi2024sam2segmentimages} and determined its orientation through a custom feature detection module that analyzed structural differences between the upper and lower parts of the object, as shown in~\Cref{fig:oru_opening_detection}. The system then decided whether the object should be flipped to ensure correct orientation for placement.
    

    \paragraph{Pin Localization} \label{par:oru_pin}
    The pin localization process was based on the estimation module from FoundationPose~\cite{wen2024foundationpose}. This framework utilized an RGB-D image along with a mesh model of the plastic base and two pins to estimate the pose of the object. 
    To improve accuracy, they incorporated a post-processing step that evaluated the quality of the estimated pose. This step relied on a simple registration error metric, which measured the sum of distances between mesh vertices and their nearest neighbors in the point cloud with a lower sum distance indicating a better alignment. 
    To refine pose estimation, the system selected the estimate with the lower registration error between the proposal with initial orientation and a proposal rotated by 180° around the vertical axis.

\subsubsection{Grasping System}
The team used a custom-designed gripper and a Cartesian impedance controller for manipulation~\cite{impedance_contr}.
The gripper featured a curved front end that centered the object and aligned its axis with the tool tip, facilitating accurate placement (\Cref{fig:oru_setup}).
The Cartesian impedance controller provided compliant behavior during motion execution in response to external forces, improving accuracy during contact-rich operations such as insertion reducing the impact of minor pose errors.

\subsubsection{Task Execution}

A Behavior Tree (BT)~\cite{bts_book} was used to define and execute the robot’s behavior, structured into three main sub-trees (\Cref{fig:oru_bt}): \textit{Detect Rack}, \textit{Picking}, and \textit{Placing}, each handling a key phase of the manipulation task.

The Detect Rack sub-tree guided the arm to the placing area and performed the pin localization~\ref{par:oru_pin}, providing a reference for accurate object placement.

The Picking sub-tree began with \textit{Detect objects in the box}, which moved the arm near the bin and ran the object detection routine~\ref{par:oru_detection}. 
This behavior included adaptive end-effector adjustments to improve detection in cluttered conditions. 
Upon successful detection, the system transitioned to \textit{Grasp object}, using the gripper's position sensor to differentiate beakers from bottles based on diameter.

The Placing sub-tree included the \textit{Detect object orientation} step~\ref{par:oru_opening}, followed by the placement action, aligning the object with the correct dishwasher pin.

To increase robustness, the BT incorporated two recovery mechanisms: one for failed grasps, using finger feedback to retry detection and grasping; and another during insertion, using force sensing to detect incorrect orientation, reorient the object, and retry placement. 
This modular structure enabled reliable task execution while handling common errors in a cluttered lab environment.

\subsubsection{Limitations}
The proposed approach made the following assumptions: (i) the bin contained only a single layer of objects, (ii) the environment was relatively uncluttered, and (iii) the system did not have a recovery mechanism if the placement failed. Potential solutions to overcome these limitations included aligning the estimated depth to metric depth, incorporating Vision-Language Models (VLM) to detect the plastic bin and dishwasher pins in a cluttered environment and to generate masks for subsequent processing, and designing a robust placement detection module using a VLM to verify the correct placement of objects.

\subsection{Politecnico di Milano (PoliMi)}

\begin{figure}[t]
    \centering
    \includegraphics[width=0.9\columnwidth]{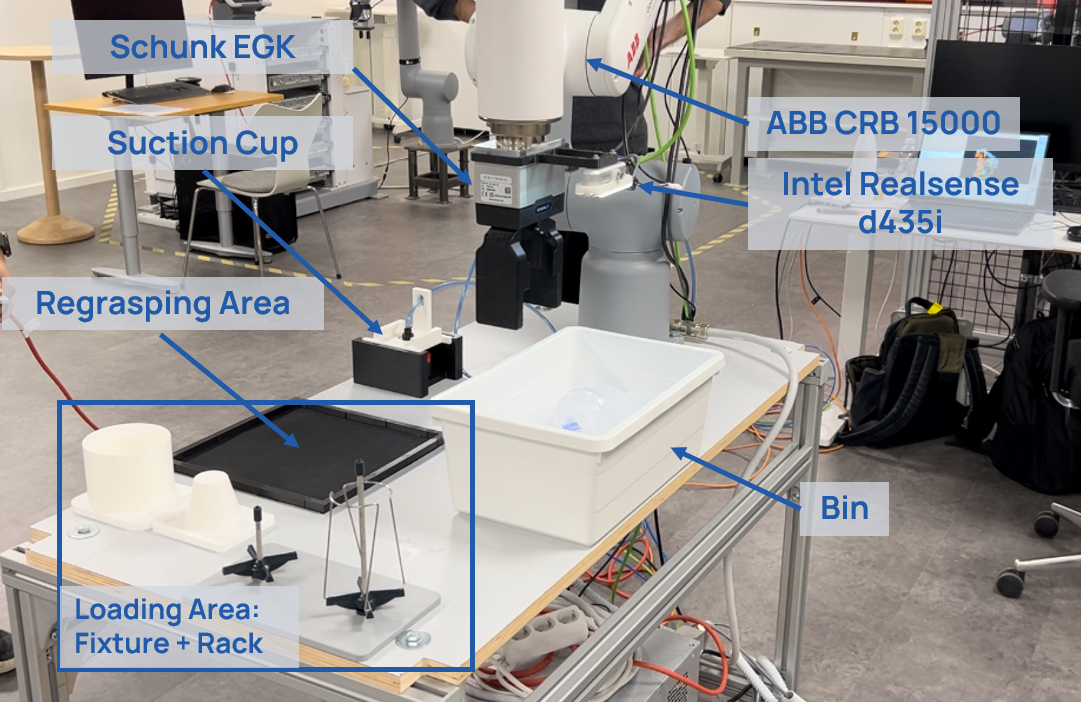}
    \caption{The setup proposed by PoliMi was built around an ABB GoFa~5 collaborative robot, equipped with a Realsense D435i stereo camera and a custom parallel gripper optimized for the pin insertion task. A detachable suction cup was also employed for the bin-picking task.}
    \label{fig:polimi_setup}
\end{figure}

\begin{figure}[t]
  \centering
  \includegraphics[width = 0.8\linewidth]{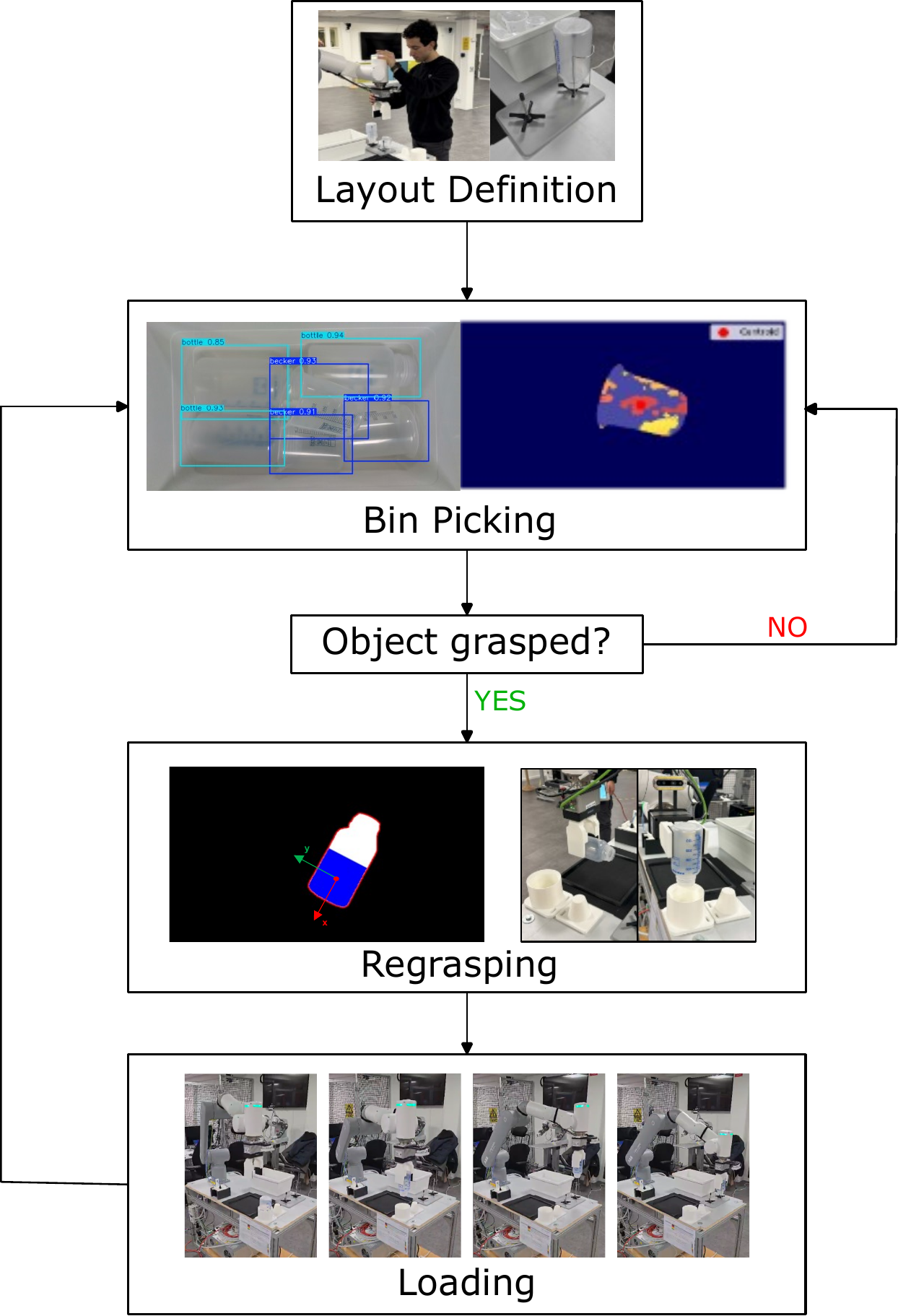}
  \caption{Team PoliMi approach involved teaching the desired tray layout through single-shot kinesthetic demonstration. Bin-picking and regrasping actions were repeated iteratively until the dishwasher tray was fully loaded.}
  \label{fig:polimi_flowchart}
\end{figure}

The PoliMi team focused on subtask B, dishwasher loading. Their setup was built around an ABB GoFa~5 equipped with a gripper featuring custom 3D-printed fingers, shown in~\Cref{fig:polimi_setup}. The gripper was mounted on a flange holding an RGB-D Realsense d435i camera. The setup also comprised a suction cup which was used to extract bottles and beakers from the bin to overcome the limitations of unfeasible grasping poses due to the finger encumbrance.
The team proposed to leverage the flexibility of a one-shot kinesthetic teaching approach, which they used to teach dishwasher loading to the robot. 
Moreover, they used visual perception to detect the desired target during bin picking and to estimate object poses for a regrasping action, which was used to position the parts on the fixtures before the loading phase. Their framework consisted of four principal components, as shown in~\Cref{fig:polimi_flowchart} and detailed below.

\subsubsection{Layout definition}
This initial step allowed the user to specify the number of bottles and beakers to be loaded into the dishwasher, along with their layout on the tray. This information was provided to the system by leveraging Programming by Demonstration. To define a new layout, the user first selected the type of item to be loaded. Then, they hand-guided the robot from the item's dedicated fixture to the dishwasher pins, storing movement waypoints and actuating the gripper when needed through the GoFa arm-side interface. Thus, the entire layout 
was defined as a set of loading skills~\cite{lucci2024intuitive}. A dedicated user interface allowed the user to easily teach and store a new layout or load a previously defined one.

\subsubsection{Bin-picking}
Given a predefined layout, 
the bin-picking phase was used to extract from the bin one of the items required by the layout. For this purpose, the robot grabbed the suction cup from its holder with the gripper and moved above the bin-picking area to allow a proper view of the entire bin with the in-hand camera. A fine-tuned YOLO-based Deep Neural Network (DNN) model~\cite{redmon2016you} recognized objects in the bin as bottles or beakers. Then, a SAM2~\cite{ravi2024sam2segmentimages} module segmented the items and estimated their picking pose. For each localized object, the depth was extracted by averaging the perceived depth over the mask. Among the objects required by the predefined layout, the closest one to the camera was selected for grasping, to avoid occlusion and facilitating extraction. Then, the picked item was released in a regrasping area in front of the robot.

\subsubsection{Regrasping}
The extracted object was prepared to be placed in the tray through a regrasping pipeline. A fine-tuned YOLO-based DNN model was used to detect the object in the grasping area that matched either the ``bottle'' or ``beaker'' label, selected during bin-picking. 
The region of interest obtained from a successful detection was fed to a SAM2 module that extracted the object mask, which was then refined through erosion and dilation steps. These morphological transformations mitigated inaccuracies arising from object transparency, which might have led to poor segmentation performances. Finally, a template-matching procedure determined the mask centroid. A candidate, top-down, grasping pose was computed by shifting this point along the principal axis of the item, identified by fitting a minimum area aligned bounding box to the mask, towards the bottom side of the object. The computed grasping pose was used to correctly grasp the object with the parallel gripper and place it in the corresponding fixture. 

\subsubsection{Loading}
Once the item was regrasped and placed in the fixture, the system loaded the waypoint list corresponding to the action being taught by the user. Thus, through simple linear waypoint movements and gripper actuation, the robot was able to load the object onto the target pin. The system then updated the item list by removing the object that had just been loaded. After that, it iterated back to the bin-picking phase until every item in the layout was loaded.

\vspace{.5cm}
The four modules were used to build the pipeline shown in~\Cref{fig:polimi_flowchart}. The initial layout definition was used to define the loading task, stating which item to place on each dishwasher pin. The loading actions composing the layout were saved as list of waypoints, taking the item from the corresponding loading area (shown in~\Cref{fig:polimi_setup}) to the desired pin. During the loading phase, these actions were executed as fixed linear waypoints motions to move the item from the initial fixture to the target pin. 
This approach enabled reuse of the layout across dishwasher cycles without relying on complex vision systems. It ensured precise and safe loading, allowing users to easily define robot movements in cluttered environments. In contrast, using vision to detect pins and loading areas on real dishwashers would have been challenging due to complex tray layouts, reflective surfaces, and moisture—factors that significantly increase perception errors.
Regarding the intermediate regrasping step, its purpose was twofold: it was used to prepare the object, with its base facing upwards, for the loading phase, but also to check whether the grasp executed during the bin-picking phase was successful or not. If no graspable object was detected, the bin-picking module was repeated until an object of the desired type was extracted and placed on the table.\\ 
Overall, the proposed solution showed reliable performance on the challenge day, being able to load both pins on every run. The cycle time for the full pipeline took around two minutes per object, from bin-picking to loading. 
Despite the heavily reflective objects, the bin-picking pipeline also showcased good performance, as image post-processing through different color filters and depth averaging managed to return feasible target poses for the suction cup.\\
The main drawback of the method was its reliance on static fixtures in the regrasping area, required as an intermediate step to complete the loading procedure. Another element that penalized cycle times was the choice of a six-degrees-of-freedom (DoF) robot, as having no redundancy translated into longer unwinding motions to prevent reaching joint limits or cable entanglement
The main failures stemmed from the suction cup getting stuck in bin corners or struggling with cluttered layers of mixed objects—though none occurred on the challenge day. Future setups should integrate the suction cup into the gripper and eliminate regrasping fixtures to enhance dexterity and flexibility.

\subsection{Royal Institute of Technology (KTH)}

The KTH team used the ABB Mobile YuMi research platform, customized multipurpose grippers, and a ROS2-based perception and navigation software stack. 
Their solution addressed the complete WARA Robotics Mobile Manipulation Challenge, thus tackling both subtasks. 

The approach made several operational assumptions to limit the task complexity.
It assumed that the approximate initial position of the cart, the positions for offloading, and the dishwasher locations were predefined. Regarding manipulation, it assumed that cups and bottles were initially placed upright on surfaces of known height that ideally had distinct, non-reflective, and non-white appearances to facilitate reliable visual detection. 
Additionally, tables and trays were assumed to be visually distinguishable.

\subsubsection{Hardware Setup}
The ABB Mobile YuMi research platform was chosen to address the challenge. It comprises an ABB Dual-Arm YuMi robot mounted on a telescopic pillar fixed on a custom-made omnidirectional base. The robotic base uses four independently steerable and drivable wheels to achieve almost-instantaneous arbitrary change in planar velocity, i.e. directional and rotational. The mobile base is equipped with two 2D lidar sensors (front and back) for mapping and localization. The manipulator is a dual-arm ABB YuMi robot. Each arm has 7-DoF with a payload of 500g. An Azure Kinect RGB-D camera, mounted atop the robot, provides visual and depth information required for perception tasks.

\begin{figure}[t]
    \centering
    \includegraphics[width=.9\linewidth]{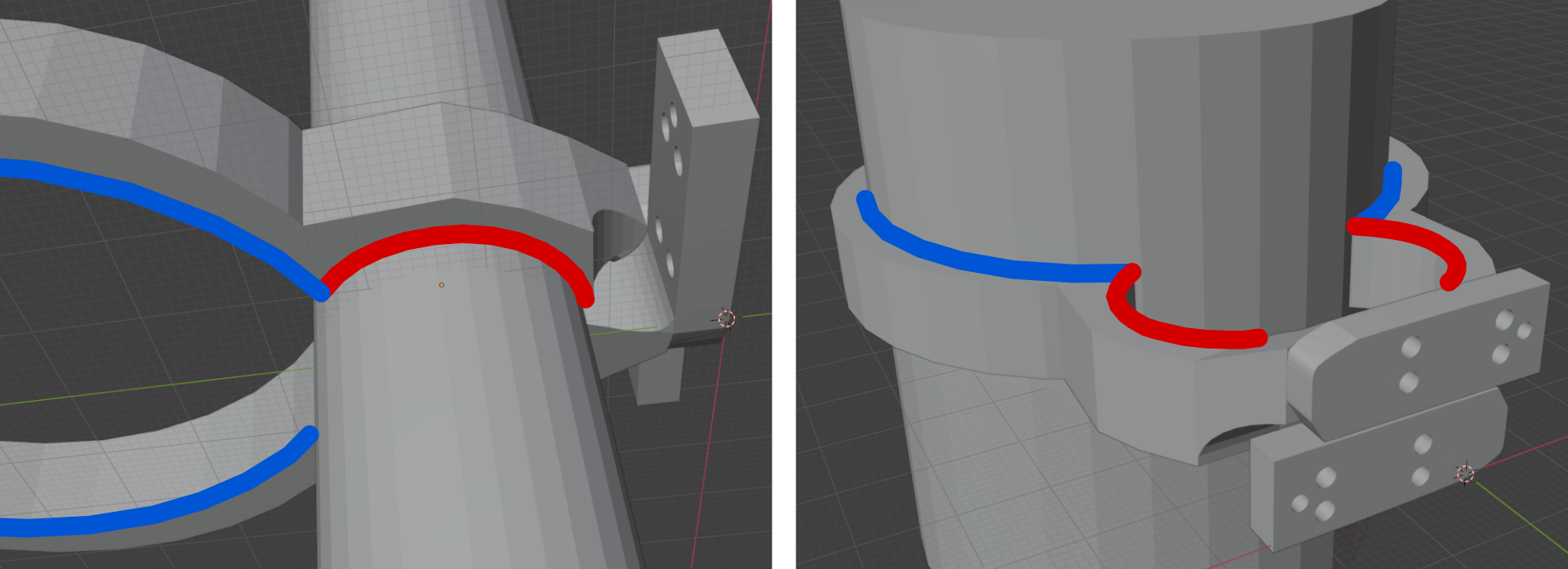}
    \caption{The KTH multi-purpose finger design is shown by the active regions, in blue for plasticware manipulation and red for cart handling.}
    \label{fig:kth-grippers}
\end{figure}

The team mounted custom-designed grippers (\Cref{fig:kth-grippers}) that offered compliant and adaptable grasping capabilities suitable for grasping of both the cart handle and cylindrical plasticware. This multi-purpose gripper design has demonstrated its effectiveness in manipulation with multiple modalities~\cite{kang2019design, aqib2024design}. 
As the gripper closed, the handle or the plasticware naturally slid to tightly match the inner surface of the fingers despite the positional uncertainty, exemplifying a mechanical funnel~\cite{mason1985mechanics}. It also showcased the usage of pregrasp cages as waypoints for object grasping~\cite{rodriguez2012caging, dong2024quasi}.

\subsubsection{Software Stack}
The software architecture utilized ROS~2 and integrated existing open-source packages with custom modules. 
Navigation functionalities were provided by SLAM Toolbox for environment mapping, Nav2 with Adaptive Monte Carlo Localization (AMCL)~\cite{zhang2009self}, and Smac Hybrid-A* planner~\cite{macenski2024open} for path generation. 
The team implemented an adaptive robot footprint that dynamically accommodated the additional geometry of the cart during transport.

The object-detection module employed a YOLO-based~\cite{redmon2016you} DNN to recognize and localize bottles and cups in the RGB images. Bounding boxes were projected into the depth space to compute the three-dimensional positions on a table or within a workspace. Plane segmentation was then applied to localize the height of objects. For dishwasher trays, a separate detection pipeline using a transformer-based model (Grounding DINO~\cite{liu2024grounding}) located black plates (the tray surface) and pins. Point-cloud processing through principal component analysis (PCA) and convex hull optimization identified the tray corners and pin positions.

The robot control was realized through high-level primitives interfacing with ABB hardware controllers through ABBs \textit{RWS} API~\cite{rws} with an adapted version of the interface from~\cite{styrud2024automatic}. An impedance control strategy ensured robustness against positional uncertainties and alignment inaccuracies during grasping and manipulation.

\subsubsection{Task-Specific Strategies}

\begin{figure}
    \centering
    \includegraphics[width=.9\linewidth]{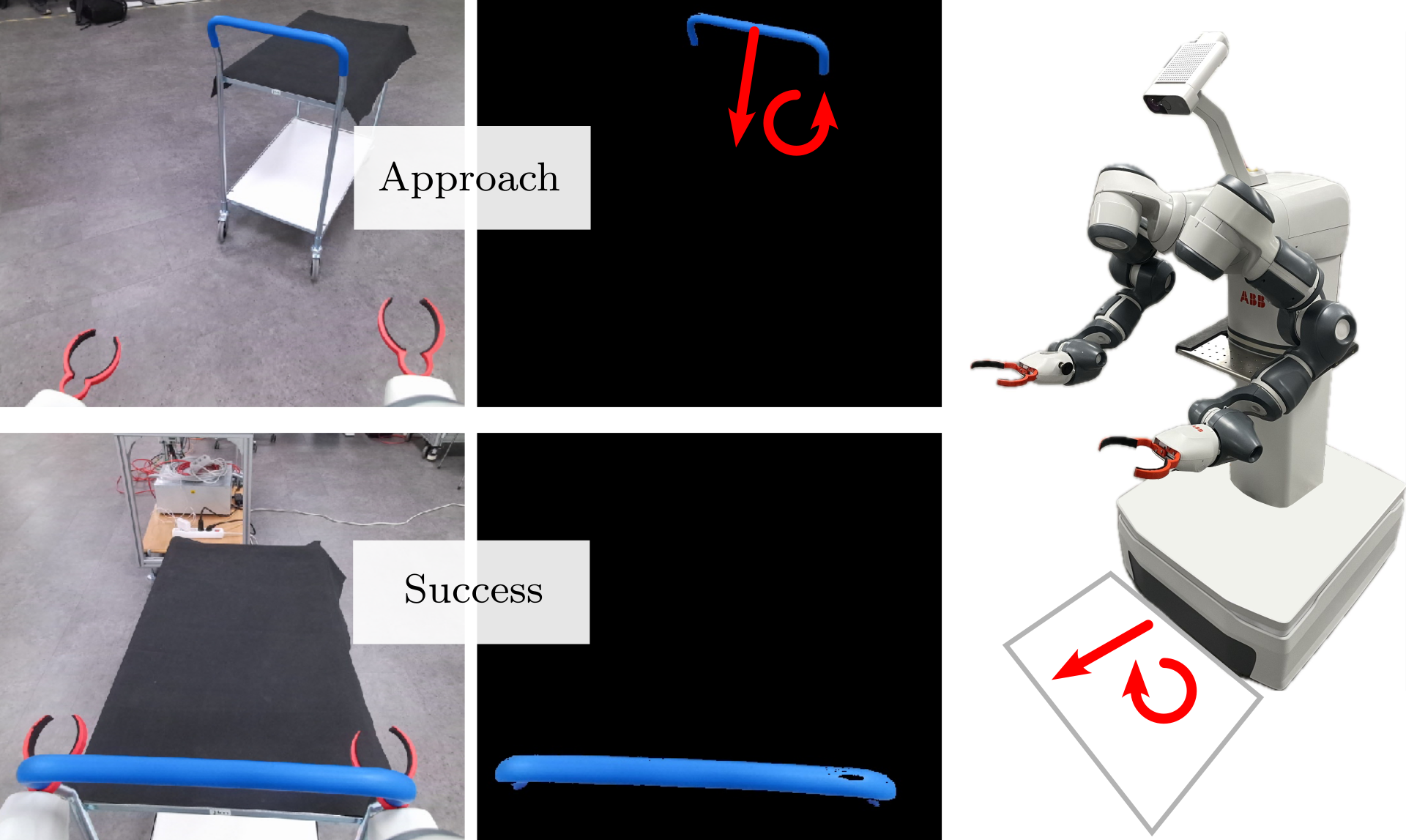}
    \caption{In KTH's carting method, the mobile base was velocity controlled and followed the illustrated color-based visual servoing policies. The impedance control of the robot arms enabled robust grasping and transportation.}
    \label{fig:kth-cart}
\end{figure}

The cart handling procedure used visual servoing based on HSV color segmentation for alignment with cart handles, followed by impedance-controlled lateral grasping. After grasping, the navigation module dynamically updated the robot's collision footprint to safely transport the cart.
The masking and visual servoing approach are shown through real examples in~\Cref{fig:kth-cart}.

The plasticware manipulation subtasks involved detecting upright cups and bottles using YOLO-based detection, computing suitable grasp poses, and executing impedance-controlled grasps. 
Placement into dishwasher trays leveraged transformer-based detection to identify tray structures and pins, with impedance-controlled insertion motions, ideally ensuring safe and precise object placement.

\subsubsection{Results and Discussion}
\begin{figure}
    \centering
    \includegraphics[width=.9\linewidth]{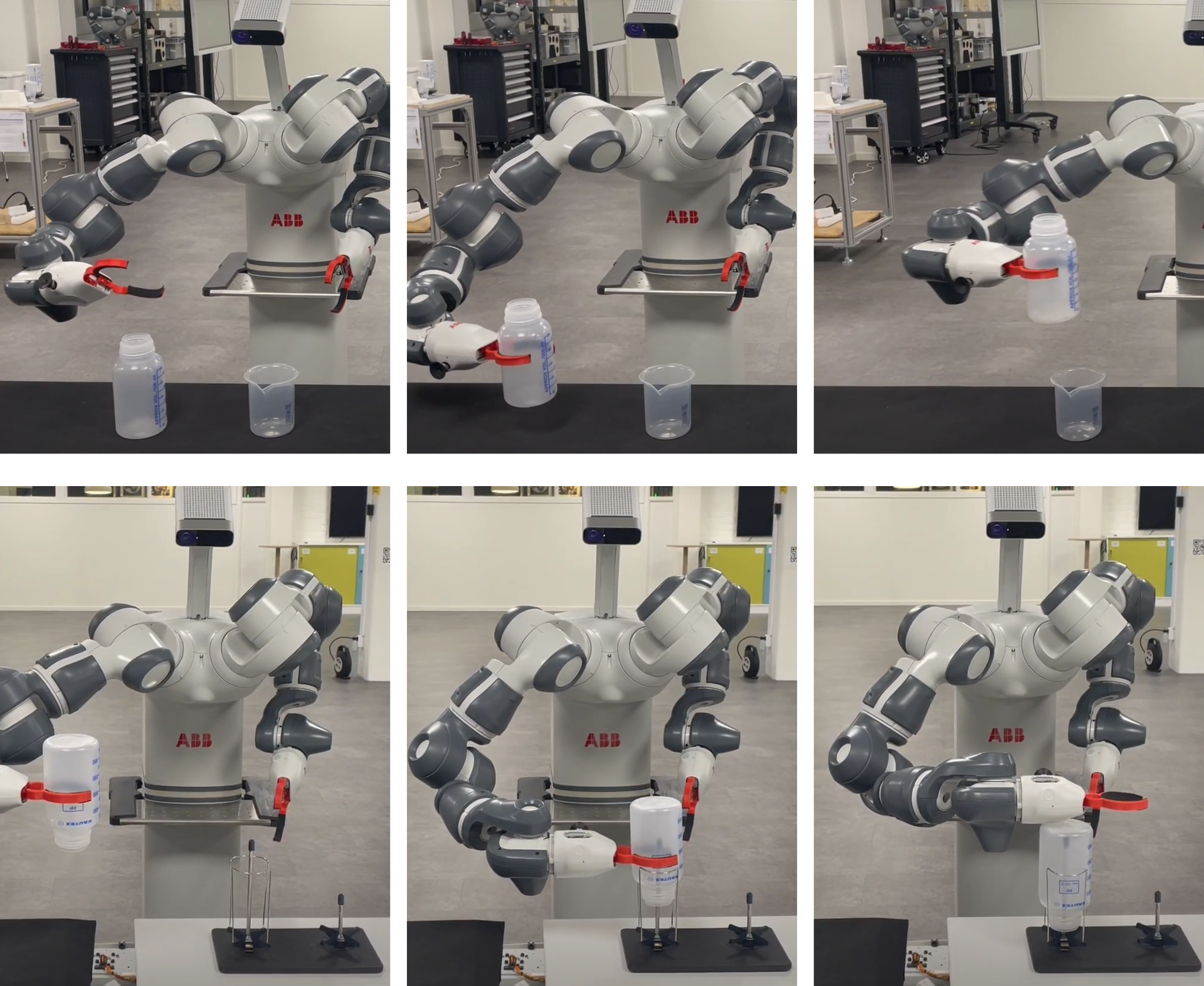}
    \caption{The KTH plasticware manipulation is shown through an example sequence. The top row shows pick procedure through the bottle-detection, grasping, and moving steps. The bottom row shows the placing procedure through the pin-detection, insertion, and compliant pushing steps.}
    \label{fig:kth-manip}
\end{figure}

Subsystem evaluations demonstrated reliable performance, although full integration between navigation and manipulation subsystems was not completed within the available timeframe. 
Plasticware detection was qualitatively accurate, and multiple successful grasps and placements were observed during experimental trials.
An example successful trial is shown as a step sequence in~\Cref{fig:kth-manip}.
Navigation and cart pushing tasks achieved reliable performance, validating the effectiveness of visual servoing and adaptive footprint planning. 
An example successful cart approach and grasp is shown in~\Cref{fig:kth-cart}. 
Overall, the described approach showcased a high degree of modularity, where off-the-shelf algorithms were integrated with custom gripper hardware and task primitives. 

While the color-based cart detection operated effectively in controlled lighting conditions, it may require more robust segmentation methods for environments with similarly colored backgrounds. Furthermore, the dishwasher-loading pipeline relied on known and consistent object orientation, limiting its applicability to objects lying on their sides or in cluttered bins. Further potential sources of failure were (i) the absence of explicit failure detection and recovery mechanisms, (ii) errors in object and goal pose estimation from the perception pipeline, and (iii) variability in wireless network communication quality with the mobile robot. 

Future iterations of the manipulation challenge could be centered around the encountered limitations and sources of failure, leading to more reliable and widely applicable solutions. 
The original plan in this project was to incorporate the skills to solve each sub-task into a reactive Behavior Tree policy~\cite{iovino2022survey}. 
While that was not possible within this project due to time constraints, it would be an interesting extension that should notably improve robustness.

\subsection{Lund Technical University (LTH)}

The LTH team focused on the autonomous cart navigation subtask A, particularly cart pulling. Their approach was to use Model-Predictive Control (MPC) to achieve holistic control of the ABB Mobile YuMi research platform. To reduce the complexity of the problem, some initial assumptions were made, which then got slightly relaxed: the robot's, cart's, and goal positions were known and accurate and the cart's handlebar was easily recognizable in the environment. 

\subsubsection{Hardware Setup}
The ABB Mobile YuMi research platform was chosen for the cart navigation task with standard off-the-shelf ABB fingers mounted on both grippers, as the gripping width was enough to hold the cart by its handlebar. 


\subsubsection{Software Stack}
The actual control of the robot was performed using ROS~2, with hardware specific drivers for each sub-robot (i.e. the torso, the pillar, and the base) and a whole-body control interface for the entirety of the robot. The control signals were generated by a custom-made MPC using the Crocoddyl optimal control library~\cite{crocoddyl}. A set of Cartesian trajectory tracking controllers were used as a fallback strategy in case of convergence issues.

\subsubsection{Modeling}
From a path-following perspective, both the dual-arm mobile manipulator and the cart were modeled as points in the $xy$-plane with a heading direction, expressed by two reference frames, $\Sigma_B$ and $\Sigma_C$ respectively, having the $z$-axis orthogonal to the world plane, expressed by the frame $\Sigma_W$. 
It was assumed that the robot was rigidly grasping the cart at its handlebar with both arms, in a symmetric fashion. In this way, $\Sigma_C$ was constrained to be between $\Sigma_L$ and $\Sigma_R$, making the cart navigation problem an absolute-relative control task. 
An additional assumption was that the mobile base was constrained to non-holonomic movements, while the cart can move freely in space.

\subsubsection{Method}
\begin{figure}[t]
    \centering
    \includegraphics[width=0.8\linewidth]{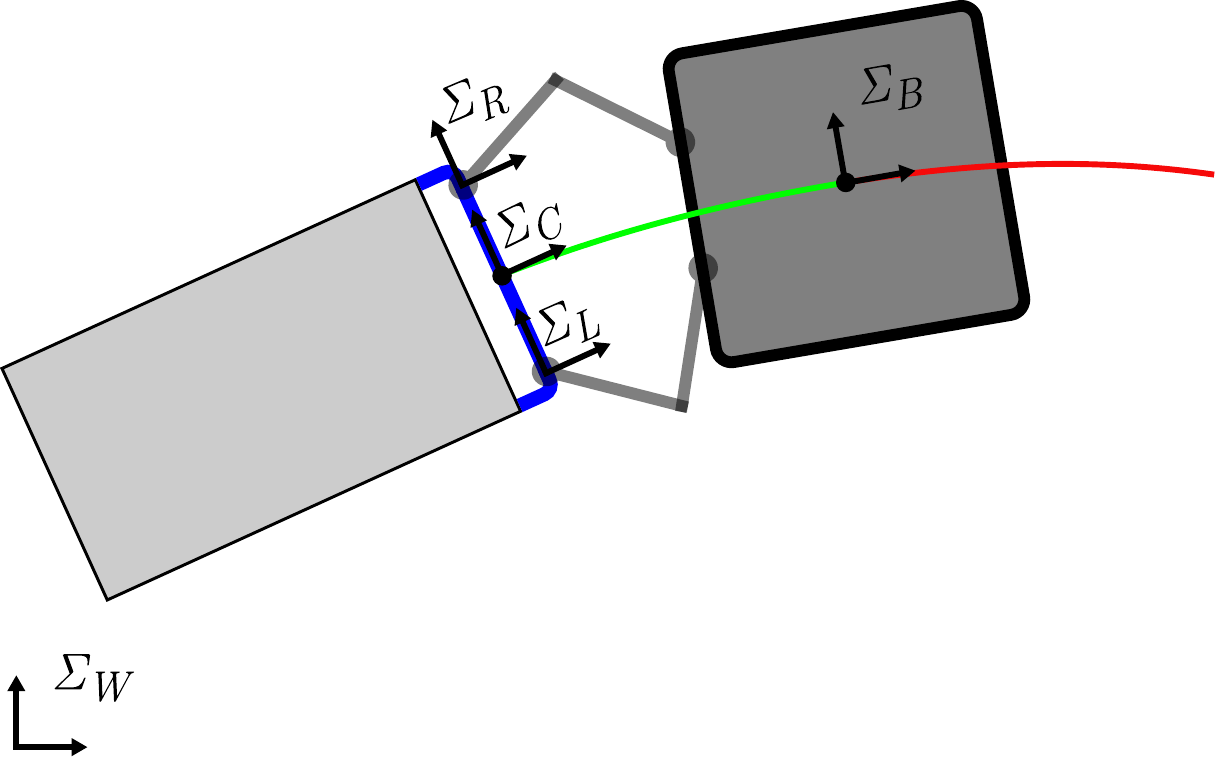}
    \caption{Model schematic. The red line represents mobile base's obstacle-free path obtained from the path planner, while the green line represents the history of base's positions, which is converted into a reference trajectory for the handlebar and thus the cart.}
    \label{fig:myumi_and_model}
\end{figure}

The cart-pulling navigation problem was split into two separate subproblems. Firstly, a smooth, safe, online path was generated for the mobile base. This path was followed by a trajectory-tracking controller moving the base along the path. Secondly, the cart was steered so as to avoid obstacles while minimizing the arms' effort and maintaining manipulability. These goals were accomplished by prescribing the cart's path to be the path traversed by the base, i.e. the path comprised of previous base positions. This way the cart also followed a smooth obstacle-free path. The model with the constructed references can be seen in~\Cref{fig:myumi_and_model}.

\paragraph{Path generation}
To navigate the cart-robot system in the environment, an obstacle-free path was generated. This was achieved in an online fashion, to account for dynamic obstacles. Before creating a path, obstacles were identified and converted to a useful representation for the planner. Lidar scanners were employed to collect distance measurements of the surrounding environment. The obtained distances were converted to a point cloud and morphologically inflated into circles in the $xy$-plane. A morphological union was then performed so to gather circles in bigger blobs, which were then simplified in the interest of computation time for the next step. The \textit{starrification} algorithm in~\cite{dahlin} was used to convert the obstacles in \textit{starshaped} ones. The obstacles were further inflated with a clearance radius using~\cite{dahlin} to create a \textit{tube-path} that guaranteed convergence to the goal pose using the planner in~\cite{huber}. The clearance radius was set to be the biggest radius that individually encircled the robot and cart. The path was a limited-horizon orbit of a dynamical system describing the flow towards the goal:
\[
\dot{r} = M(r, \mathcal{O}) (r^g - r)
\]
with $r$ the current robot position, $r^g$ the desired goal, $M(\cdot,\cdot)$ a matrix that describes the attractive dynamics, and $\mathcal{O}$ is the set of world obstacles.

\begin{figure}[t]
    \centering
    \includegraphics[trim=30 30 30 30,clip,width=0.240\linewidth]{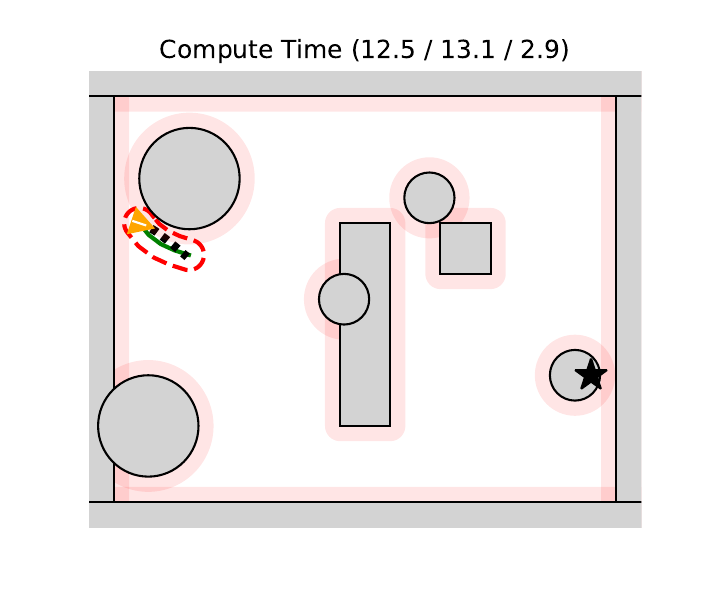}
    \hfill
    \includegraphics[trim=30 30 30 30,clip,width=0.240\linewidth]{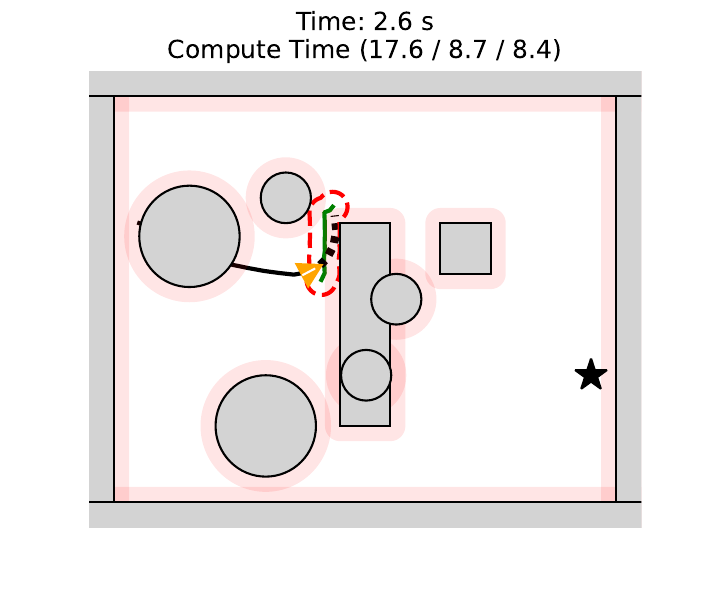}
    \hfill
    \includegraphics[trim=30 30 30 30,clip,width=0.240\linewidth]{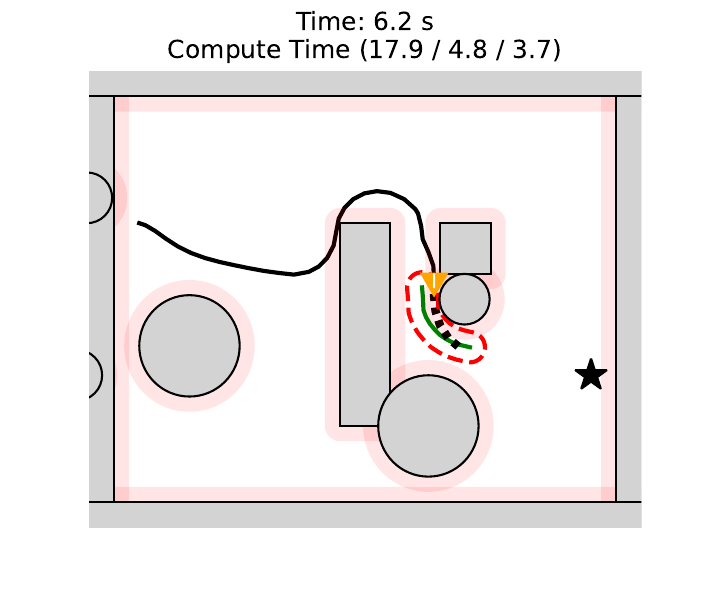}
    \hfill
    \includegraphics[trim=30 30 30 30,clip,width=0.240\linewidth]{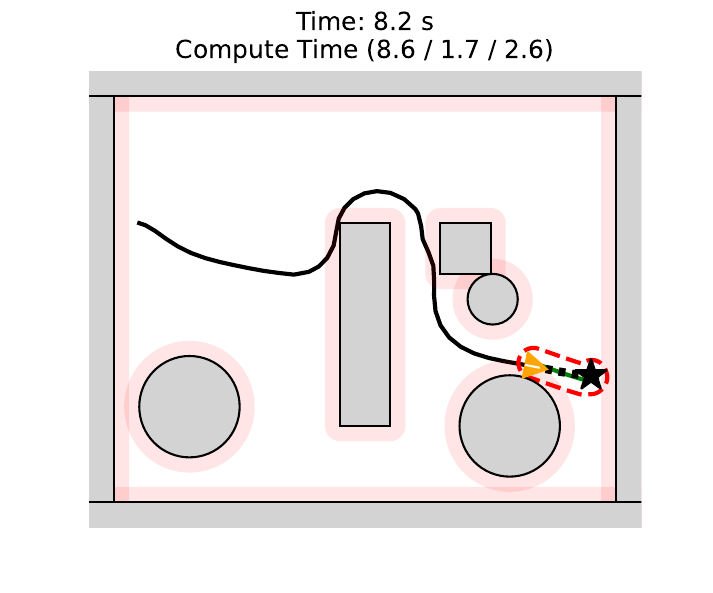}
    \caption{Path generation for a unicycle robot with dynamic obstacles and safety guarantees over time (left to right). The robot (yellow arrow) follows the kinematically constrained trajectory (black dashed) while staying inside the safe area (red boundary) of the local desired trajectory (green solid), reaching the goal (star), never colliding with the dynamic obstacles (circles, translating vertically or horizontally). The overall motion is depicted in solid black, while the obstacle safety boundaries are depicted in light red. Generated with~\cite{dahlin}.}
    \label{fig:path_generation}
\end{figure}

\paragraph{Base path following}
Given a path, the robot's mobile base was tasked with moving backward while minimizing the time required to reach the goal. An MPC controller was employed to move the robot on track in the shortest time. The same controller was also tasked with arm control.
The prescribed path consisted of both positions and orientations.
%
%

\paragraph{Cart trajectory following}
The $\textit{SE(2)}$ path for the cart was constructed from the $\textit{SE(2)}$ path previously traversed by the
base. Due to the cart being a rigid body and the rigid grasping of the cart's handlebar, the cart's path was transformed into $\textit{SE(3)}$ paths for the left and the right gripper. The $(x,y)$ translations were provided by base's past path, as well as one degree of rotation. The height of the handlebar was fixed, and the rigid grasp covered the remaining 2 degrees of rotation.  To avoid extending the arms and avoid collision between the cart and the base of the robot,
the cart's path started at a predetermined fixed arc-length distance between the base's current position and a point among its previously traversed positions. Finally, the same timing law imposed on the base was used for the end-effector paths.


Upon grasping of the handlebar, the initial path between cart and robot was set to be a Bezier curve that smoothly connects the cart with the base. This path was subsequently updated with base's positions.

\subsubsection{Results}
The implementation was not finished in time for the challenge day, on which simulation results were shown within the presentation of the team's approach. However, the proposed solution was pursued further, with additional lab visits, for the purpose of a standalone publication which is in progress at the time of writing.


\section{Conclusions and Lessons Learned}

The first WARA Robotics Mobile Manipulation Challenge proposed a highly unstructured lab automation use-case and gave the freedom to the participating teams to choose the robotic platform as well as the solution strategy. A video of the challenge day is available online\footnote{\url{https://youtu.be/F9PRgmFeArM?si=dyfY8xrQoy5UWdw6}}.

The teams from Politecnico di Milano (PoliMi) and Örebro University (ÖrU) tackled the manipulation sub-task (Section~\ref{sec:rules}-\textit{B}) with an ABB GoFa~5 and a Franka Emika Panda robot respectively.
The solution from PoliMi achieved the highest robustness both in the perception and manipulation components. With the former they were able to handle the case where the bin was fully loaded with objects laying in multiple layers, while in the latter they could generalize to a more complex dishwasher tray system (e.g., featuring more pins). However, their approach relied on a more complicated grasping solution as well as fixtures. 
On the other hand, the solution from ÖrU could generalize to slight variations on the position of the bin and the dishwasher tray, and they traded off a simpler and more polished robotic system with a slight decay on the success rate. However, by representing the whole task plan policy as a Behavior Tree, they could recover to failures in the perception and grasping without halting the robot execution.

The team from Lund Technical University (LTH) focused instead to the carting sub-task (Section~\ref{sec:rules}-\textit{A}), using the ABB Mobile YuMi research platform. They formulated the problem as cart pulling instead of pushing and used a whole-body control approach. The method was effectively working in simulation but the team was not able to fully integrate it on the real platform by the challenge day.

Finally, the team from the Royal Institute of Technology (KTH) was the only one to attempt solving the full challenge task. Although they had working solutions for both sub-tasks individually, they were not able to combine them in a single execution. Moreover, their approach for the manipulation sub-task required bottles and beakers to be standing upright instead of lying down in the bin.

A jury composed of employees from ABB and AstraZeneca was tasked to evaluate the teams' performances and named Örebro University as the winner team to receive the first prize: a 50.000 SEK participation bonus to the 2025 Automatica\footnote{\url{https://automatica-munich.com/en/trade-fair/}} Exhibition in Munich.

The target of this first edition of the WARA Robotics Mobile Manipulation Challenge was to raise awareness in the academic sector about the WARA Robotics lab in Västerås. This environment is dedicated to facilitate experimental validation of academic research in the field of robotics and AI through a tight connection with industrial partners who provide technical resources as well as challenging industrial use-cases.



The challenge also provided valuable lessons in how to design effective robotics competitions. Flexibility in the choice of the platform and problem interpretation allowed creativity, but also made direct comparison between solutions difficult. A more standardized setup\----including a shared robotic platform, digital twin environment, and clear evaluation metrics\----will be introduced in the next edition to support reproducibility of the results and transparency and fairness on the evaluation.

In summary, the challenge not only showcased promising approaches to real-world mobile manipulation, but also highlighted key areas where academic solutions still struggle with robustness and scalability. It also reinforced the value of using challenge-based formats to assess the technology readiness level (TRL) of academic research in industry-relevant contexts. These insights will help guide future challenges and foster stronger collaboration between academia and industry.


\bibliographystyle{ieeetran}
\bibliography{references.bib}

\end{document}